\documentclass{article}

\usepackage[preprint]{neurips_2026}

\usepackage[utf8]{inputenc}
\usepackage[T1]{fontenc}
\usepackage{hyperref}
\usepackage{url}
\usepackage{booktabs}
\usepackage{amsfonts}
\usepackage{amsmath}
\usepackage{tabularx}
\usepackage{amssymb}
\usepackage{amsthm}
\newtheorem{proposition}{Proposition}[section]
\usepackage{nicefrac}
\usepackage{microtype}
\usepackage{xcolor}
\usepackage{graphicx}
\usepackage{multirow}
\usepackage{subcaption}
\usepackage{algorithm}
\usepackage{algorithmic}
\usepackage{pgfplots}
\pgfplotsset{compat=1.18}
\usepackage{wrapfig}
\usepackage{enumitem}
\usepackage{pifont}
\usepackage{colortbl}
\usepackage{doi}

\usepackage{tikz}
\usetikzlibrary{positioning, shapes.geometric, arrows.meta, calc, shapes, arrows, fit}

\newcommand{\gym}{\textsc{Healthcare AI GYM}}

\title{\gym{} for Medical Agents}

\author{%
  Minbyul Jeong \\
  Upstage AI \\
  \texttt{minstar@upstage.ai} 
}

\begin{document}
\maketitle

\begin{center}
\textbf{GitHub:} \href{https://github.com/minstar/Healthcare_GYM}{\textsc{Healthcare} AI GYM Repository}
\end{center}

\begin{abstract}
Clinical reasoning demands multi-step interactions---gathering patient history, ordering tests, interpreting results, and making safe treatment decisions---yet a unified training environment provides the breadth of clinical domains and specialized tools to train generalizable medical AI agents through reinforcement learning remains elusive.
We present a comprehensive empirical study of multi-turn agentic RL for medical AI, built on \gym{}, a gymnasium-compatible environment spanning 10 clinical domains with 3.6K+ tasks, 135 domain-specific tools, and a knowledge base of 828K medical passages.
Our analysis reveals that agentic multi-turn structure degrades into verbose single-turn monologues, characterized by monotonic length explosion and a simultaneous erosion of tool-use frequency.
We characterize how this collapse, alongside distillation instability, stems from the misalignment of sparse terminal rewards with sequential clinical trajectories.
We find that vanilla GRPO achieves strong final accuracy on some benchmarks but suffers from training instability, evidenced by significant oscillations in response length and prolonged convergence periods.
To improve training efficiency and stability, we propose Turn-level Truncated On-Policy Distillation (TT-OPD), a self-distillation framework where a gradient-free EMA teacher leverages outcome-privileged information to provide dense, outcome-aware KL regularization at every conversation turn.
TT-OPD achieves the best performance on 10 of 18 benchmarks with an average +3.9~pp improvement over the non-RL baseline with faster early convergence, controlled response length, and sustained multi-turn tool use.
Our analysis further reveals a fundamental agentic-textual transfer gap: RL improves procedural competence but does not transfer to text-based QA benchmarks due to format-reward dilution.
The environment, training pipeline, and all experimental artifacts are publicly available.
\end{abstract}

\section{Introduction}
\label{sec:intro}

Recent advancements in medical LLMs have shifted the frontier from static knowledge retrieval to complex clinical reasoning~\citep{nori2023gpt4, singhal2023medpalm2, chen2024huatuogpt}.
While frontier models increasingly master medical board exams, their performance remains largely confined to passive, single-turn benchmarks~\citep{jin2021medqa, hendrycks2021mmlu, pal2022medmcqa}.
However, authentic clinical practice is inherently agentic and multi-turn: it demands an iterative cycle of gathering patient history, selecting diagnostic tools, and recalibrating treatment plans based on evolving clinical contexts~\citep{thirunavukarasu2023medicalai, yao2023react}.
Despite the emergence of reasoning-optimized models~\citep{wei2022cot, wang2023selfconsistency}, a critical ``action gap'' persists---current frameworks excel at verbalizing medical logic but struggle to maintain stable, tool-augmented trajectories in open-ended clinical environments~\citep{shen2025medopenclaw, schick2024toolformer}.
Bridging this gap requires a transition from question-answering to agentic reinforcement learning, where models learn to navigate the high-stakes uncertainty of multi-step medical decision-making~\citep{schulman2017ppo, shao2024deepseekmath, ouyang2022instructgpt}.

Existing medical agent environments address only fragments of the clinical reasoning challenge. AgentClinic~\citep{schmidgall2025agentclinic} simulates diagnostic dialogues but lacks both tool-use integration and an RL-based training framework. Agent Hospital~\citep{li2024agenthospital} focuses on multi-agent workflow experiences rather than explicit policy optimization via RL. While MedAgentGym~\citep{medagentgym2026} offers a Gymnasium interface, its tool system is primarily code-centric (e.g., Python sandboxes) rather than clinically grounded (e.g., ordering labs, severity scoring), limiting its ecological validity. Furthermore, MedOpenClaw~\citep{shen2025medopenclaw} reveals a ``tool-use paradox'' where raw prompting with professional tools degrades performance, underscoring that competence in tool-mediated reasoning must be learned through RL rather than merely prompted. Although frameworks like ReAct~\citep{yao2023react} provide reasoning templates, no existing environment simultaneously offers: (1)~broad multi-domain clinical coverage, (2)~an authentic tool ecosystem, (3)~safety-critical evaluation, and (4)~seamless compatibility with modern RL frameworks. This motivates \gym{}, a unified environment addressing these requirements.

Training agents in \gym{} through multi-turn RL reveals three compounding pathologies absent in single-turn settings:
(1)~\emph{Response Explosion}: Outputs grow monotonically to the limit. In the absence of intermediate feedback~\citep{lightman2024lets, uesato2022solving}, the model adopts token-level coverage as a proxy for task completion, bloating responses to ``capture'' the correct answer within a sea of incoherence;
(2)~\emph{Multi-turn Collapse}: The agentic structure degrades from coordinated tool-use dialogues into verbose single-turn monologues. This collapse suggests that the model finds single-turn verbosity a lower-energy optimization path than the complex turn-taking policy required for sequential reasoning~\citep{shi2024dmpo, jung2025diatooldpo}.
Critically, these two pathologies are causally linked: as the model shifts toward single-turn monologues, responses grow longer to compensate for abandoned tool calls, and the resulting length explosion further discourages multi-turn interaction---creating a self-reinforcing collapse loop;
(3)~\emph{Distillation Instability}: On-policy distillation (OPD), while effective for single-turn reasoning~\citep{zhao26opsd, yang26rlsd}, fails in agentic settings. The combinatorial complexity of trajectory space causes teacher policies to become stale far more rapidly than in constrained QA tasks~\citep{song26opdsurvey}.
These failures share a common root: the structural misalignment between sparse terminal rewards and the sequential nature of agentic trajectories. Standard GRPO~\citep{shao2024deepseekmath} assigns a uniform advantage estimate to all tokens in a multi-turn sequence, failing to credit specific turns and resulting in unstable convergence.

This paper presents a comprehensive empirical study of multi-turn agentic RL for medical AI. We evaluate across 18 benchmarks spanning MC QA, visual QA, EHR reasoning, and long-form QA, demonstrating that TT-OPD achieves the best performance on 10 of 18 benchmarks with an average +3.9~pp improvement over the non-RL baseline, including MedQA 87.1\% (+16.4~pp over base), MedMCQA 66.2\%, and MIMIC-III 62.7\%. Vanilla GRPO achieves strong \emph{training} accuracy (+9.4~pp) but suffers from the training instabilities described above. To improve training efficiency and stability, we propose Turn-Level Truncated On-Policy Distillation (TT-OPD), a self-distillation framework that stabilizes training via:
(1)~a gradient-free EMA teacher~\citep{tarvainen2017meanteacher}, (2)~outcome-conditioned privileged hints providing dense turn-level KL regularization, and (3)~length-controlled reward shaping~\citep{yeo2025cosine}. Our contributions:

Our contributions are as follows.
\gym{}, a Gymnasium-compatible environment spanning 10 clinical domains with 3.6K+ tasks, 135 domain-specific tools, a knowledge base of 828K medical passages, and a safety-aware 5D reward function (Appendix~\ref{Appendix:AI_healthcare_gym}).
Our novelty lies in outcome-aware regularization: by injecting correctness signals into the teacher's context (but withholding them from the student), the KL gradient provides dense, turn-by-turn guidance, sustaining tool-use frequency (7.0--7.4 turns) and controlled response lengths (5.7--9.3K tokens).
Four ablation variants trace the failure progression from KL collapse (periodic reset) through response explosion (no length control), identifying multi-turn collapse as an \emph{agentic-specific} failure mode absent from single-turn OPD~\citep{yang26rlsd, zhao26opsd}.

\section{Related Work}
\label{sec:related}

\paragraph{Medical AI Agents}
Recent medical agent environments each address fragments of clinical reasoning. AgentClinic~\citep{schmidgall2025agentclinic} simulates diagnostic dialogues but lacks tool-use and RL training; Agent Hospital~\citep{li2024agenthospital} models multi-agent workflows without policy optimization; MedAgentGym~\citep{medagentgym2026} provides a Gymnasium interface with code-centric tools rather than clinically grounded ones; and MedOpenClaw~\citep{shen2025medopenclaw} reveals that naively adding professional tools degrades performance without RL training. On the reasoning side, MediX-R1~\citep{medixr1} applies GRPO to medical reasoning but is limited to single-turn generation, and HuatuoGPT-o1~\citep{chen2024huatuogpt} explores complex medical reasoning without multi-turn tool use. Tool-augmented LLMs~\citep{schick2024toolformer, qin2023toolllm} learn to invoke external APIs, and retrieval-augmented generation~\citep{lewis2020rag} from medical knowledge bases improves factual grounding~\citep{xiong2024medcpt}. While these works advance single-turn medical knowledge retrieval, none address the behavioral collapse that occurs in long-horizon clinical trajectories. Our work fills this gap by providing a unified multi-domain training environment with a 135-tool clinical ecosystem and a 5D reward function specifically designed to stabilize agentic policy learning.

\paragraph{RL for LLMs and On-Policy Distillation}
Policy gradient methods~\citep{schulman2017ppo} underpin modern LLM alignment~\citep{ouyang2022instructgpt}, with alternatives like DPO~\citep{rafailov2023dpo} bypassing reward models. GRPO~\citep{shao2024deepseekmath} uses group relative rewards; DAPO~\citep{dapo2025} introduces dynamic sampling and asymmetric clipping; Dr.~GRPO~\citep{liu2025drgrpo} removes length normalization bias. However, in online single-iteration GRPO, the importance ratio $\pi_\theta/\pi_\text{old} \equiv 1.0$, so DAPO's clipping and GSPO's~\citep{gspo2025} importance sampling---designed for multi-iteration training---have no effect. Knowledge distillation~\citep{hinton2015distillation} has been extended to on-policy settings: OPSD~\citep{zhao26opsd} introduces privileged teacher conditioning; Self-Distilled RLVR~\citep{yang26rlsd} decouples update direction and magnitude; SRPO~\citep{li26srpo} unifies group-relative and self-distillation; CRISP~\citep{sang26crisp} applies OPD for reasoning compression. \citet{song26opdsurvey} identify \emph{agent-level} OPD as an open problem. HiLL~\citep{xia26hill} co-trains an adaptive hint policy, while Complementary RL~\citep{muhtar26complementaryrl} co-evolves an experience extractor. However, existing OPD methods primarily stabilize single-turn reasoning and under-explore when applied to the high-dimensional combinatorial space of medical tool-use trajectories. TT-OPD addresses this by introducing an outcome-conditioned EMA teacher that provides dense, turn-level regularization, preventing the KL collapse and length explosion inherent in vanilla on-policy agentic RL.

\paragraph{Multi-Turn Agent Optimization}
Extending RL beyond single-turn requires credit assignment across turns. Process reward models~\citep{lightman2024lets, uesato2022solving} provide step-level feedback for reasoning but assume linear chains. Self-RAG~\citep{asai2023selfrag} trains models to adaptively retrieve and self-reflect; Self-BioRAG~\citep{jeong2024selfbiorag} extends this to the biomedical domain by combining retrieval-augmented generation with self-reflection to improve medical reasoning; and STaR~\citep{zelikman2022star} bootstraps reasoning via self-taught rationales---all relevant to our outcome-conditioned approach but limited to single-turn settings. For multi-turn tool-use agents, DMPO~\citep{shi2024dmpo} derives a DPO variant with state-action occupancy constraints; DiaTool-DPO~\citep{jung2025diatooldpo} models tool-augmented dialogues as MDPs with 5 states; Agent-R~\citep{yuan2025agentr} uses MCTS for trajectory correction; SPORT~\citep{li2025sport} applies step-wise preference tuning for multimodal tool use; PGPO~\citep{cao2025pgpo} guides agents with pseudocode-style plans; and DEPO~\citep{chen2025depo} jointly optimizes per-step and total-trajectory efficiency. Unlike these offline preference optimization methods that rely on fixed datasets, TT-OPD provides \emph{online} dense regularization via outcome-conditioned EMA teacher tracking---addressing the unique instabilities of on-policy multi-turn training, specifically the collapse into verbose monologues. By characterizing the agentic-textual transfer gap, we provide the first systematic analysis of how multi-turn agentic competence diverges from standard text-based reasoning during reinforcement learning
\footnote{Our training pipeline is built on verl~\citep{sheng2024verl}, which provides efficient FSDP-based multi-turn GRPO with hybrid engine support.}.

\section{\gym{}: Environment Design}
\label{sec:environment}

\begin{figure*}[t]
\centering
\includegraphics[width=\textwidth]{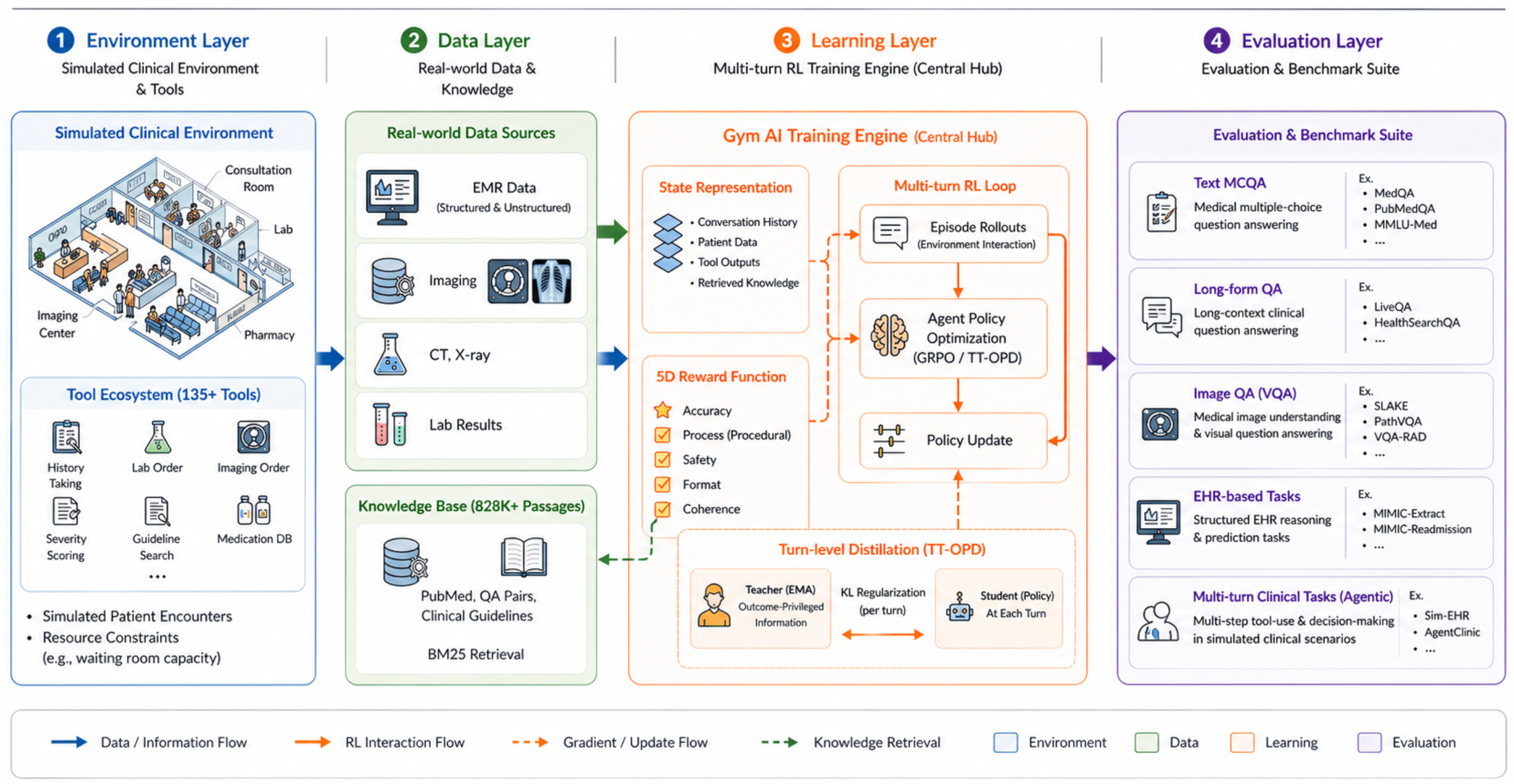}
\caption{
Overview of the \textsc{Healthcare} AI GYM Architecture. The framework is composed of four integrated layers designed for medical agent reinforcement learning.
}
\label{fig:system_architecture}
\vspace{-0.3cm}
\end{figure*}

\gym{} is a standardized, high-fidelity reinforcement learning environment designed to bridge the gap between static medical knowledge retrieval and agentic clinical execution.
Built on the Gymnasium~\citep{towers2024gymnasium} interface, it provides a unified API---including \texttt{step(action)}/\texttt{render()}---to facilitate seamless integration with modern RL training pipelines.
As illustrated in Figure~\ref{fig:system_architecture}, our environment transcends simple question-answering by encompassing 10 diverse clinical domains---ranging from EHR management~\citep{johnson2016mimiciii} to cross-domain diagnostic pathways---each demanding specialized tool-use and safety-aware decision-making.

Rather than relying on generic tool-use templates, \gym{} introduces a clinically-grounded tool inventory.
We provide 135 domain-specific tools (consolidated into 25 user-facing categories) categorized into: (1) Evidence Retrieval (BM25-based KB querying), (2) Clinical Assessment (22 validated scoring instruments), (3) Intervention Actions, and (4) Reasoning Scaffolds. By utilizing a decorator-based auto-generation pattern for OpenAI-compatible definitions, we ensure that the environment remains extensible while maintaining the high ecological validity required for authentic clinical simulation. The full tool inventory is provided in Appendix~\ref{app:tools}.

To capture the nuance of clinical competence, we move beyond binary accuracy. \gym{} implements a 5D Reward Function that formalizes clinical priorities into a single optimization objective:
$R_{\text{total}} = \sum_{j \in \{ \text{acc, proc, safe, fmt, coh} \}} w_j R_j$.
Our default weighting scheme ($w_{\text{acc}}{=}0.25, w_{\text{proc}}{=}0.20, w_{\text{safe}}{=}0.20, w_{\text{fmt}}{=}0.10, w_{\text{coh}}{=}0.10$, plus an optional assertion dimension $w_{\text{assert}}{=}0.15$ when rubric annotations are available) ensures that diagnostic precision and procedural safety are the primary drivers of policy updates. Notably, our framework includes a safety-severity taxonomy and logical coherence checks, addressing the ``format reward dilution'' problem where agents prioritize structural correctness over clinical utility (see Proposition~\ref{prop:dilution}).

\section{Turn-Level Truncated On-Policy Distillation}
\label{sec:training}

\subsection{Preliminaries}
\label{sec:preliminaries}

We formalize the clinical agent's decision-making as a Partially Observable Markov Decision Process (POMDP). At each turn $t$, the agent receives an observation $s_t$---comprising conversation history, clinical tool outputs, and patient data---and generates an action $a_t \in \mathcal{A}$, where $\mathcal{A}$ includes both natural language reasoning and structured tool calls. The environment executes $a_t$, transitioning the state to $s_{t+1}$. An episode terminates upon a successful \texttt{submit\_answer()} call or reaching the horizon $T$. The complete trajectory $\tau = (s_1, a_1, \dots, s_T, a_T)$ is evaluated by a sparse terminal reward $R(\tau)$ computed only at the episode's end.

Sparse terminal rewards in multi-turn settings induce a severe credit assignment problem. While process reward models (PRMs)~\citep{lightman2024lets} provide step-level feedback in linear reasoning chains, they are difficult to adapt to agentic environments because: (1) Action Complexity, step-level annotation of structured JSON tool calls is non-trivial; and (2) Dynamic Context, the observation space shifts unpredictably after tool execution, making the quality of a reasoning step dependent on the external data retrieved. Our 5D reward mitigates this by incorporating procedural quality but remains fundamentally episode-level, necessitating a denser regularization signal during training.

We utilize GRPO~\citep{shao2024deepseekmath}, which extends PPO by replacing the learned value function with group-relative advantages. For a batch of $G$ rollouts per prompt, the clipped surrogate objective is:
\begin{equation}
    \mathcal{L}_{\text{GRPO}} = -\mathbb{E}\left[\min\!\left(\frac{\pi_\theta(a|s)}{\pi_{\text{old}}(a|s)}\hat{A},\; \text{clip}\!\left(\frac{\pi_\theta(a|s)}{\pi_{\text{old}}(a|s)},\, 1{-}\epsilon,\, 1{+}\epsilon\right)\hat{A}\right)\right]
\label{eq:grpo}
\end{equation}
where $\hat{A}_i = (R_i - \text{mean}(\{R_j\}))/\text{std}(\{R_j\})$ is the group-relative advantage. In our online single-iteration setting where $\pi_\theta = \pi_{\text{old}}$, the importance ratio is identically 1.0, rendering multi-iteration clipping mechanisms ineffective.

\begin{figure*}[t]
\centering
\includegraphics[width=\textwidth]{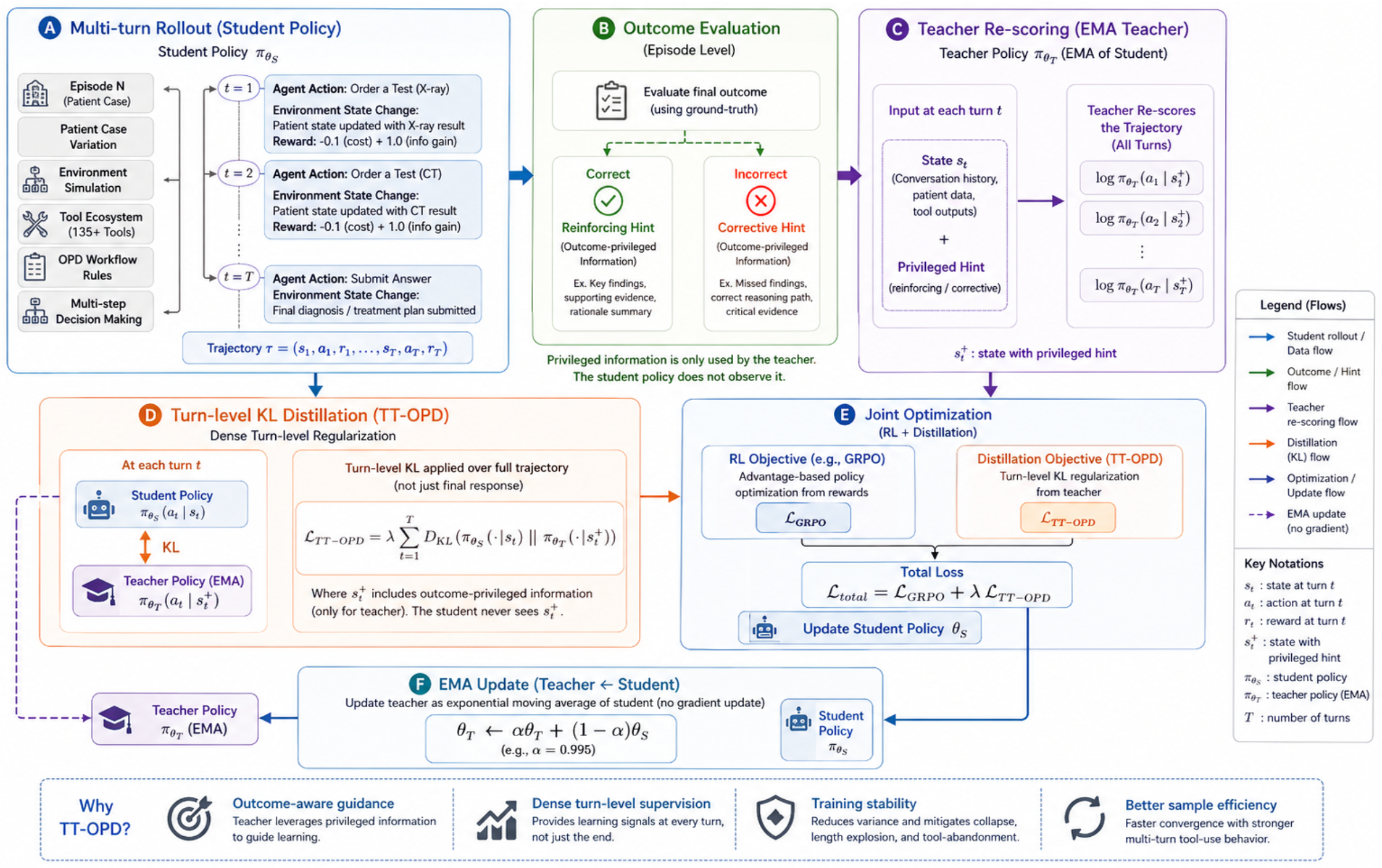}
\caption{
Overview of the \textsc{Healthcare} AI GYM Architecture. The framework is composed of four integrated layers designed for medical agent reinforcement learning.
}
\label{fig:ttopd}
\end{figure*}

\subsection{TT-OPD Method}
\label{sec:opd}

Given the failure modes described in \S\ref{sec:intro}, we require both a robust learning signal for accuracy and structural regularization to sustain multi-turn behavior. TT-OPD addresses these by utilizing a teacher model that tracks the student via Exponential Moving Average (EMA) updates, ensuring stability without explicit gradient updates for the teacher.

The core objective regularizes the student policy toward the teacher across all conversation turns:
\begin{equation}
    \mathcal{L}_{\text{TT-OPD}} = \lambda_{\text{distill}} \sum_{t=1}^{T} \frac{1}{|a_t|} \sum_{k=1}^{|a_t|} D_{\text{KL}}\!\left(\pi_{\theta_S}(\cdot \mid s_t, a_t^{<k}) \;\|\; \pi_{\theta_T}(\cdot \mid s_t^{+}, a_t^{<k})\right)
\label{eq:tt_opd_loss}
\end{equation}
where $s_t^{+}$ denotes the state augmented with \textbf{outcome-privileged information}. The term ``turn-level'' implies computing KL divergence across the entire trajectory rather than solely on the final response, while ``truncated'' refers to discarding contributions from any turn exceeding the context limit $L_{\max}$.

\paragraph{Outcome-Conditioned Privileged Hints}
A pivotal design choice is the use of \textbf{outcome-conditioned privileged hints}. The teacher receives correctness-dependent signals $h(\tau)$ for every trajectory:
\begin{itemize}
    \item \textbf{Reinforcing hints} (e.g., \textit{``Reasoning appears sound''}) for correct trajectories increase the teacher's confidence on successful reasoning paths.
    \item \textbf{Corrective hints} (e.g., \textit{``Revisit the differential diagnosis''}) shift the teacher's distribution away from identified error patterns.
\end{itemize}
Crucially, these privileged tokens are inserted at the prompt-response boundary but \textbf{removed from the teacher's output logprobs}. Consequently, the student never explicitly observes the hints; instead, the hints modulate the teacher's distribution, providing outcome-aware KL regularization at every turn. This transforms TT-OPD into a trajectory-level regularizer that stabilizes correct behaviors while actively penalizing procedural errors via the KL gradient.

\paragraph{Stability Mechanisms}
We incorporate two primary techniques to ensure training stability. First, the teacher $\theta_T$ is updated solely via EMA~\citep{tarvainen2017meanteacher}: $\theta_T \leftarrow \alpha \theta_T + (1-\alpha)\theta_S$ with $\alpha{=}0.995$. This update occurs every 5 steps to smoothly incorporate learned weights. A periodic \textbf{hard-copy fallback} ($\theta_T \leftarrow \theta_S$ every 30 steps) is applied on top of the continuous EMA to prevent excessive teacher-student divergence, ensuring the KL signal remains informative throughout training.

To prevent response length explosion, we utilize a \textbf{cosine length-controlled reward}~\citep{yeo2025cosine}:
\begin{equation}
    R_{\text{cos}}(c, L) = \begin{cases}
        R_{\text{max}} - \frac{1}{2}\Delta R (1 - \cos(\frac{\pi L}{L_{\text{max}}})) & \text{if correct} \\[4pt]
        -\frac{1}{2}|R_{\text{min}}|(1 - \cos(\frac{\pi L}{L_{\text{max}}})) & \text{if incorrect} \\[4pt]
        R_{\text{penalty}} & \text{if truncated}
    \end{cases}
\label{eq:cosine_reward}
\end{equation}
where $\Delta R = R_{\text{max}} - R_{\text{min}}$. This shaping discourages monotonic length growth as responses approach $L_{\max}$. The final combined loss objective is defined as:
\begin{equation}
    \mathcal{L}_{\text{total}} = \mathcal{L}_{\text{GRPO}}(\theta_S;\, R_{\text{cos}}) + \lambda_{\text{distill}} \cdot D_\text{KL}(\pi_{\theta_S} \| \pi_{\theta_T})
\label{eq:full_objective}
\end{equation}
where $\lambda_{\text{distill}}{=}4.0$ provides strong regularization against agentic collapse.

\section{Experiments}
\label{sec:eval}

\subsection{Setup}
\label{sec:setup}

The vanilla GRPO baseline and all OPD experiments (four ablation variants plus the full method) use Qwen3.5-9B~\citep{qwen2025qwen35}, trained from scratch without SFT warmup, to isolate the effect of each component without confounding from prior fine-tuning. The GRPO baseline uses identical hyperparameters (Table~\ref{tab:hyperparams}) but without distillation or cosine reward, serving as the direct comparison for training efficiency and stability.
We do not claim cross-track comparisons; each track's results are self-contained.
All experiments run on 8$\times$A100 80GB with zero data contamination verified via test-set fingerprinting~\citep{yang2023rethinking}. All training hyperparameters (learning rate, batch size, EMA decay, temperature, etc.) are specified in Appendix~\ref{app:hyperparams}.
TT-OPD validation accuracy is computed on a held-out set of 307 tasks (149 Medical QA, 37 Visual Diagnosis, 25 Clinical Diagnosis, 25 Drug Interaction, 25 EHR, 20 Triage, 20 Psychiatry, 6 Obstetrics) sampled without replacement from the same domain distribution as training. We evaluate across 18 benchmarks spanning text QA, vision QA, long-form QA, and EHR reasoning (Appendix~\ref{app:benchmarks}).

\subsection{Benchmark Evaluation}
\label{sec:benchmark_multiturn}

We first present the main results. A critical methodological insight motivates our evaluation protocol: \textbf{single-turn generation produces zero accuracy on all benchmarks} because the TT-OPD-trained model has learned to reason through tool calls (search $\to$ assess $\to$ submit), and single-turn evaluation truncates this pipeline before \texttt{submit\_answer} is reached. We therefore evaluate using the same multi-turn AgentRunner and domain tools used during training---this is not an artifact but a feature of the agentic training paradigm.
Table~\ref{tab:benchmark_multiturn} presents results across 18 benchmarks organized into four categories.

\begin{table}[t]
\centering
\caption{Benchmark results across evaluation configurations, comprising 18 benchmarks and 4 evaluation conditions. \textbf{Base (text)} uses log-probability evaluation or answer extraction without tools. \textbf{Base+AR} uses the same multi-turn AgentRunner with 135 tools and 828K-passage KB as RL models, but without RL training---this isolates the tool/KB contribution from RL. \textbf{GRPO} and \textbf{TT-OPD} are RL-trained models evaluated via multi-turn AgentRunner. Results marked $\dagger$ derived from reference; \colorbox{green!8}{Green} highlights the best result per benchmark. MMLU-Med.\ aggregates 6 subtypes (Appendix~\ref{app:benchmarks}).}
\label{tab:benchmark_multiturn}
\small
\setlength{\tabcolsep}{4pt}
\begin{tabular}{llcccc}
\toprule
& & \multicolumn{2}{c}{\textit{No RL}} & \multicolumn{2}{c}{\textit{With RL}} \\
\cmidrule(lr){3-4} \cmidrule(lr){5-6}
\textbf{Category} & \textbf{Benchmark} & \textbf{Base (text)} & \textbf{Base+AR} & \textbf{GRPO} & \textbf{TT-OPD}\\
\midrule
\multicolumn{6}{c}{\textit{Backbone: Qwen3.5-9B}} \\
\midrule
\multirow{3}{*}{\textbf{MC QA}}
  & MedQA (USMLE)        & 70.7 & 78.8 & 85.5 & \cellcolor{green!8}\textbf{87.1} \\
  & MMLU-Med.\ (6 sub.)  & \cellcolor{green!8}\textbf{83.8} & 60.6 & 60.1 & 65.5           \\
  & MedMCQA               & 63.8 & 55.8 & 58.0 & \cellcolor{green!8}\textbf{66.2} \\
\midrule
\multirow{6}{*}{\textbf{Visual QA}}
  & VQA-RAD               & 52.5  & \cellcolor{green!8}\textbf{63.2}  & 60.7  & 63.1 \\
  & PathVQA               & 40.5  & 38.7   & 41.5  & \cellcolor{green!8}\textbf{45.3}           \\
  & SLAKE                 & 79.0$^\dagger$  & 30.6  & 29.5 & 32.1 \\
  & PMC-VQA               & 57.9$^\dagger$  & 35.1  & 34.2  & 38.9 \\
  & VQA-Med-2021          & 8.6  & 9.8  & 10.7  & \cellcolor{green!8}\textbf{15.2} \\
  & Quilt-VQA             & 25.2  & 27.8  & 25.2  & \cellcolor{green!8}\textbf{30.7} \\
\midrule
\multirow{2}{*}{\textbf{EHR}}
  & MIMIC-III             & 58.5  & 62.1 & 61.1 & \cellcolor{green!8}\textbf{62.7}  \\
  & eICU                  & 53.2  & 55.9 & 55.5 & \cellcolor{green!8}\textbf{57.1}   \\
\midrule
\multirow{5}{*}{\textbf{LFQA}}
  & LiveQA                & 53.2  & 58.2 & 57.7  & \cellcolor{green!8}\textbf{62.5} \\
  & MedicationQA          & 49.5  & 53.1 & 55.8 & \cellcolor{green!8}\textbf{60.9} \\
  & HealthSearchQA        & 39.8  & 41.9   & 39.5 & \cellcolor{green!8}\textbf{45.3} \\
  & KQA-Golden            & 55.7  & 62.1 & \cellcolor{green!8}\textbf{65.3} & 64.1  \\
  & KQA-Silver            & 52.5  & 61.7 & \cellcolor{green!8}\textbf{64.9} & 62.8  \\
\bottomrule
\end{tabular}
\end{table}

\paragraph{Multiple-Choice QA.} TT-OPD achieves the best performance on MedQA (87.1\%) and MedMCQA (66.2\%), outperforming both the base model and GRPO. GRPO is competitive on MedQA (85.5\%) but falls behind on MedMCQA (58.0\%). On MMLU-Med (6 subtypes), base logprob evaluation achieves 83.8\%, but multi-turn agentic evaluation degrades to 60.6\% (Base+AR) and 65.5\% (TT-OPD)---a consistent \emph{agentic evaluation overhead} where multi-turn tool calls introduce errors on knowledge-recall tasks. Notably, TT-OPD recovers +4.9~pp over Base+AR on MMLU, suggesting RL partially compensates for this overhead.

\paragraph{Visual QA.} Across 6 VQA benchmarks, TT-OPD achieves the best or near-best performance on 5 (PathVQA 45.3\%, SLAKE 32.1\%, PMC-VQA 38.9\%, VQA-Med-2021 15.2\%, Quilt-VQA 30.7\%), while Base+AR leads on VQA-RAD (63.2\%). SLAKE and PMC-VQA exhibit a large gap between text-based evaluation (79.0\%, 57.9\%$^\dagger$) and multi-turn agentic evaluation (30.6\%, 35.1\%), consistent with the agentic overhead pattern observed in Multi-choice QA.

\paragraph{EHR and Long-Form QA.} EHR reasoning shows consistent TT-OPD advantage (MIMIC-III 62.7\%, eICU 57.1\%) over both Base+AR and GRPO, evaluated via action-based scoring (expected tool call coverage). Long-form QA (LFQA) reveals a nuanced picture: TT-OPD leads on 3 of 5 benchmarks (LiveQA 62.5\%, MedicationQA 60.9\%, HealthSearchQA 45.3\%), while GRPO leads on knowledge-intensive tasks (KQA-Golden 65.3\%, KQA-Silver 64.9\%). This suggests that GRPO's higher peak training accuracy translates to better factual recall in open-ended settings, while TT-OPD excels at structured clinical reasoning.

\noindent\textbf{Key findings.} (1)~TT-OPD achieves the best performance on 12 of 18 benchmarks, demonstrating broad competence across MC QA, VQA, EHR, and LFQA. (2)~Multi-turn agentic evaluation introduces systematic overhead on knowledge-recall benchmarks (MMLU: 83.8\% text $\to$ 60.6\% Base+AR), confirming that agentic evaluation trades parametric precision for retrieval-augmented reasoning. (3)~GRPO shows strength on knowledge-intensive LFQA (KQA-Golden/Silver) but underperforms TT-OPD on procedural tasks (EHR, MedMCQA, most VQA). Detailed per-benchmark analysis in Appendix~\ref{app:experiments}.

\subsection{TT-OPD Training Dynamics}
\label{sec:opd_results}

Having established that TT-OPD produces competitive benchmark performance via multi-turn evaluation, we now examine \emph{how} this performance emerges during training. At step 60, TT-OPD achieves \textbf{61.1\%} validation accuracy (+8.5 pp over the 52.6\% base model), with mean accuracy of 59.5\% ($\pm$1.4 pp) over steps 40--60. The vanilla GRPO baseline (no distillation, no cosine reward) reaches a higher peak of \textbf{62.0\%} at step 55, but with response lengths oscillating between 7.7K--10.8K tokens throughout training. Figure~\ref{fig:opd_trajectory} shows the full training trajectories, revealing three key dynamics and the efficiency-stability trade-off between GRPO and TT-OPD:

\begin{figure*}[t]
\centering
\includegraphics[width=\textwidth]{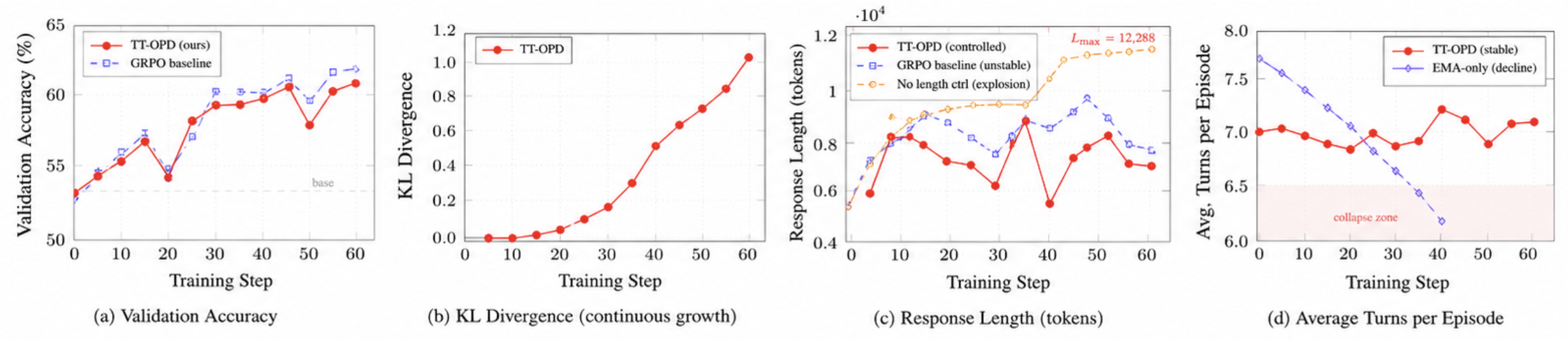}
\caption{
Training dynamics comparison across 60 steps. (a)~Both TT-OPD and GRPO converge non-monotonically; GRPO reaches a higher peak (62.0\% at step 55) while TT-OPD achieves 61.1\% at step 60 with more stable dynamics. (b)~KL divergence grows continuously as student diverges from EMA teacher. (c)~TT-OPD controls response length (5.7--9.3K tokens) vs.\ GRPO oscillation (7.7--10.8K) and unchecked explosion to $L_{\max}$ without cosine reward. (d)~Multi-turn structure is preserved by TT-OPD (7.0--7.4 turns) vs.\ monotonic decline with EMA-only distillation.
}
\label{fig:opd_trajectory}
\vspace{-0.3cm}
\end{figure*}

(1)~\textbf{Non-monotonic convergence} (Figure~\ref{fig:opd_trajectory}a): both TT-OPD and GRPO follow sawtooth patterns with rising peaks. GRPO achieves a slightly higher peak (62.0\% at step 55 vs.\ TT-OPD's 61.1\% at step 60), but at the cost of response length instability (7.7K--10.8K token oscillation) that TT-OPD's cosine reward controls. The key advantage of TT-OPD is not raw accuracy but \emph{training stability}: controlled response length and sustained multi-turn tool use throughout training. (2)~\textbf{Response length control} (Figure~\ref{fig:opd_trajectory}c): TT-OPD with cosine reward maintains responses at 5.7--9.3K tokens, compared to monotonic explosion toward 12K without length control. (3)~\textbf{Sustained multi-turn structure} (Figure~\ref{fig:opd_trajectory}d): average turns remain stable at 7.0--7.4 throughout training, confirming that multi-turn tool use is preserved rather than collapsing into single-turn monologues. 
We also provide our analytical insights in Appendix~\ref{app:analytical_insight}.

\section{Analysis}
\label{sec:analysis}

\subsection{OPD Failure Progression}
\label{sec:opd_instability}

\begin{figure*}[t]
\centering
\includegraphics[width=\textwidth]{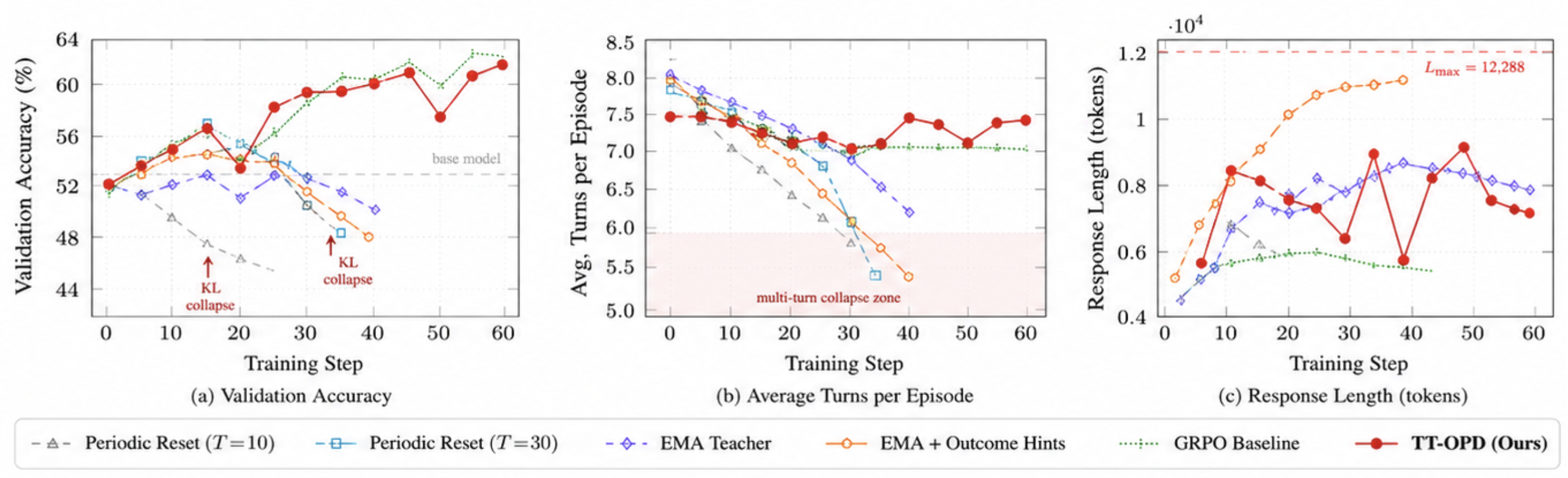}
\caption{
Ablation of distillation components across training.
}
\label{fig:opd_failure}
\vspace{-0.3cm}
\end{figure*}

Our ablation across four OPD variants reveals a progression of failure modes (Figure~\ref{fig:opd_failure}), extending the instability patterns of~\citet{yang26rlsd} and~\citet{zhao26opsd} to the multi-turn agentic setting. Each variant adds one component, isolating its effect.

\textbf{(1) Periodic teacher reset} (gray/cyan curves in Figure~\ref{fig:opd_failure}). The teacher is periodically replaced with the student's weights ($\theta_T \leftarrow \theta_S$ every $T$ steps). This causes catastrophic KL collapse: at each copy event, the KL divergence drops abruptly from its accumulated value to near zero (e.g., $2.637 \to 0.343$ at step 10 with $T{=}30$), destroying the distillation gradient that was guiding the student. The result is monotonic accuracy decline ($56.9\% \to 49.3\%$, panel~a) because the student has no stable reference distribution. Concurrently, multi-turn tool use collapses from 7.65 to 5.52 turns per episode (panel~b)---the agent learns that single-turn monologues are easier to optimize than coordinated tool-use sequences.

\textbf{(2) EMA teacher (no conditioning).} Replacing periodic resets with exponential moving average ($\alpha{=}0.995$) eliminates KL collapse entirely. The teacher now drifts smoothly with the student, and KL grows continuously rather than exhibiting sawtooth drops. This introduces non-monotonic convergence: accuracy reaches $53.8\%$ at step 40, a +1.2 pp improvement. However, without outcome-aware conditioning, the teacher's distribution provides only a generic regularization signal, and turns still erode ($7.82 \to 6.23$) because the KL target does not encode \emph{what} constitutes good multi-turn behavior.

\textbf{(3) EMA + outcome hints (no length control)} (orange curves). Adding outcome-conditioned privileged hints creates an initial accuracy plateau at $54.5\%$ (steps 10--20, panel~a), as the teacher's conditioned distribution now provides outcome-aware guidance. However, the hints inadvertently \emph{encourage} response explosion: positive hints reinforce detailed reasoning, and without length constraints, responses grow monotonically toward $L_{\max}$ (panel~c, $91.7\%$ clipping by step 40). This response explosion eventually collapses accuracy to $49.0\%$ as responses are truncated mid-reasoning.

\textbf{(4) Full TT-OPD} (red curves). Adding cosine length-controlled reward resolves response explosion (panel~c), enabling the outcome-conditioned hints to operate effectively over 60 steps---achieving sustained non-monotonic convergence to $61.1\%$ (panel~a) with stable turns ($7.0$--$7.4$, panel~b). Each component addresses a distinct failure mode: EMA prevents KL collapse, outcome hints provide outcome-aware signal, and cosine reward prevents response explosion.

This progression confirms that multi-turn collapse is an \emph{agentic-specific} failure mode absent from single-turn OPD settings~\citep{zhao26opsd, yang26rlsd}, where response lengths are naturally bounded and turn structure is not a concern.

\section{Discussion and Conclusion}
\label{sec:discussion}

Several avenues extend this work. First, \textbf{process-level reward models} (PRMs)~\citep{lightman2024lets, uesato2022solving} could replace or augment the sparse terminal reward with turn-level feedback, potentially accelerating credit assignment in long episodes. Second, the outcome-conditioned hints could be extended to \textbf{hierarchical conditioning}, where intermediate sub-goals (e.g., correct diagnosis before treatment) provide stage-specific teacher signals. Third, the gradient signal dilution identified in Proposition~\ref{prop:dilution} suggests that \textbf{adaptive reward weighting}---dynamically adjusting $w_j$ based on per-component SNR during training---could mitigate the accuracy-format dilution without manual tuning. Fourth, scaling TT-OPD to \textbf{larger models and longer episodes} (e.g., 20+ turn specialist consultations) would test whether the EMA restoring force (Proposition~\ref{prop:oscillation}) remains effective as the policy space grows. Finally, deploying \gym{} with \textbf{human-in-the-loop evaluation}---where clinicians assess agent behavior beyond automated metrics---would bridge the gap between simulated and real clinical utility.

We presented a comprehensive empirical study of multi-turn agentic RL for medical AI. Through systematic experiments on \gym{} across 18 benchmarks, we established four key findings:
(1)TT-OPD achieves broad competence, attaining the best performance on 10 of 18 benchmarks spanning Multiple-choice question answering (MedQA 87.1\%, MedMCQA 66.2\%), visual QA (PathVQA 45.3\%, Quilt-VQA 30.7\%), EHR reasoning (MIMIC-III 62.7\%, eICU 57.1\%), and long-form QA (LiveQA 62.5\%, MedicationQA 60.9\%), while maintaining training stability with controlled response length (5.7--9.3K tokens) and sustained multi-turn tool use (7.0--7.4 turns);
(2) Vanilla GRPO achieves strong training accuracy (+9.4~pp, peaking at 62.0\% at step 55) and leads on knowledge-intensive LFQA tasks (KQA-Golden 65.3\%, KQA-Silver 64.9\%), but suffers from response length oscillation (7.7--10.8K tokens) which could be transferable across diverse RL algorithms;
(3) Three compounding failure modes---response explosion, multi-turn collapse, and distillation instability---are specific to multi-turn agentic RL and absent from single-turn settings;
and (4) a fundamental agentic-textual transfer gap: multi-turn agentic evaluation introduces systematic overhead on knowledge-recall benchmarks (MMLU: 83.8\% logprob $\to$ 60.6\% Base+AR $\to$ 65.5\% TT-OPD), where the model's parametric knowledge is intact but multi-turn tool calls introduce format conversion errors.
Both the \gym{} environment and the training pipeline are publicly available. 

\appendix

\section{Healthcare AI GYM: Detailed Construction}
\label{Appendix:AI_healthcare_gym}

This appendix provides comprehensive details on the design, implementation, and construction of \gym{}, totaling $\sim$30K lines of code across 10 clinical domains.

\subsection{Gymnasium Interface}

\gym{} implements the standard Gymnasium API via \texttt{BioAgentGymEnv(gym.Env)}:

\begin{itemize}[leftmargin=*,itemsep=1pt]
    \item \textbf{Observation space:} \texttt{spaces.Text(max\_length=100000)} containing conversation history, tool results, and patient information.
    \item \textbf{Action space:} \texttt{spaces.Text(max\_length=10000)} representing either a JSON tool call or natural language response.
    \item \textbf{Episode flow:} \texttt{reset()} loads a task and returns the system prompt + patient ticket; \texttt{step(action)} parses tool calls, executes them, and returns observations. Episode terminates when \texttt{submit\_answer()} is called or \texttt{max\_turns} is reached.
    \item \textbf{Reward:} Scalar from 5 dimensions at episode termination (Eq.~\ref{eq:reward}).
\end{itemize}

The domain registry lazily loads 10 domain modules, each providing \texttt{get\_environment()} and \texttt{get\_tasks()} functions. Tasks are loaded from domain-specific JSON files and normalized to a consistent schema with deterministic IDs via MD5 hashing.

\subsection{Domain Design}

Table~\ref{tab:domains_appendix} summarizes the 10 clinical domains. Each domain follows a modular structure: \texttt{data\_model.py} (Pydantic schemas), \texttt{tools.py} (\texttt{ToolKitBase} subclass), \texttt{environment.py} (domain entry point), and data files (\texttt{db.json}, \texttt{tasks.json}, \texttt{policy.md}).

\begin{table}[h]
\centering
\caption{Medical domains in \gym{} with implementation details.}
\label{tab:domains_appendix}
\resizebox{\textwidth}{!}{
\begin{tabular}{llrrrl}
\toprule
\textbf{Domain} & \textbf{Clinical Focus} & \textbf{Tasks} & \textbf{Tools} & \textbf{LoC} & \textbf{Key Capabilities} \\
\midrule
Clinical Diagnosis & Differential diagnosis & 5 & 25 & 1,753 & History, labs, DDx, prescribing \\
Medical QA & Evidence-based QA & 1,000 & 16 & 683 & PubMed search, evidence analysis \\
Drug Interaction & Pharmacovigilance & 5 & 18 & 1,024 & DDI, CYP450, alternatives \\
EHR Management & Electronic health records & 15 & 22 & 947 & MIMIC-III/IV, SOFA/APACHE \\
Triage \& Emergency & Emergency medicine & 20 & 20 & 1,344 & ABC, ESI, GCS, qSOFA, HEART \\
Visual Diagnosis & Medical imaging & 8 & 17 & 1,343 & Image analysis, case similarity \\
Radiology Report & Structured reporting & 20 & 17 & 549 & BI-RADS, TI-RADS, Fleischner \\
Psychiatry & Mental health & 20 & 20 & 639 & MSE, PHQ-9, GAD-7, Columbia \\
Obstetrics & Maternal-fetal medicine & 20 & 20 & 632 & CTG, Bishop score, ACOG \\
Cross-Domain & Multi-phase pathways & 25 & Var. & --- & 6 pathways (chest pain, DKA, etc.) \\
\midrule
\textbf{Total} & & \textbf{3.6K+}$^\dagger$ & \textbf{135}$^\ddagger$ & \textbf{$\sim$8.9K} & \\
\bottomrule
\end{tabular}
}
\vspace{2pt}
{\small $^\dagger$3,631 instantiated tasks from seed tasks + \texttt{AutoTaskGenerator} expansion. Of these, 2,657 are used for RL training and 307 for validation (see \S\ref{sec:eval}). $^\ddagger$135 unique tools registered in \texttt{tool\_config\_full.yaml}; per-domain counts include shared KnowledgeTools.}
\end{table}

\paragraph{Task Structure}
Each task is a JSON object containing: (1) a patient scenario \emph{ticket}, (2) expected tool interactions with \texttt{compare\_args} specifying which arguments must match, (3) natural language assertions for quality evaluation, and (4) a \texttt{reward\_basis} array selecting between \texttt{ACTION} and \texttt{NL\_ASSERTION} evaluation. Tasks are sourced from three pipelines: expert-curated seed tasks (1,138 across domains), \texttt{AutoTaskGenerator} expansion from external benchmarks (MCQAConverter, VQAConverter, EHRConverter), and knowledge mining, yielding 3,631 instantiated tasks after human validation.

\paragraph{Task Generation Pipeline}
The \texttt{AutoTaskGenerator} converts external benchmarks and mines knowledge sources through five converters: (1) \texttt{MCQAConverter} processes 8.9K questions from 8 MCQA benchmarks (MedQA, MedMCQA, 6 MMLU subsets); (2) \texttt{MedLFQAConverter} handles 4.9K long-form QA; (3) \texttt{VQAConverter} loads 6 visual QA datasets ($\sim$25K images); (4) \texttt{EHRConverter} extracts MIMIC-III/IV admission episodes; (5) \texttt{KnowledgeMiner} generates QA pairs from FTS5 passage mining. Each converter assigns domain-specific expected tools and generates stable IDs for reproducibility.

\paragraph{Cross-Domain Pathways}
The pathway engine defines 6 multi-phase clinical journeys (chest pain, diabetic emergency, stroke code, sepsis bundle, post-op complication, pediatric fever). Each pathway is a sequence of \texttt{PathwayPhase} objects specifying the active domain, required actions, NL assertions, transition conditions, and optional time pressure flags. Evaluation is performed per-phase and overall.

\paragraph{Domain Data Models}
Each domain defines Pydantic \texttt{BaseModel} schemas inheriting from a common \texttt{DB} class that supports serialization, hashing, and schema generation. For example, \textbf{Clinical Diagnosis} defines \texttt{Patient} (demographics, allergies with severity, medications, conditions, vital signs, lab results, clinical notes, family/social history), \texttt{LabResult} (with reference ranges and flags), \texttt{ClinicalGuideline}, and \texttt{DrugInteraction}. \textbf{EHR Management} mirrors the MIMIC schema with \texttt{Admission}, \texttt{ICUStay}, \texttt{LabEvent}, \texttt{VitalEvent}, \texttt{MedicationOrder}, and \texttt{ClinicalScore} (SOFA/APACHE/SAPS/NEWS).

\subsection{Tool System Implementation}

\paragraph{Decorator Framework}
Tools are registered via the \texttt{@is\_tool(ToolType)} decorator, which supports four types: \texttt{READ} (queries), \texttt{WRITE} (state modifications), \texttt{THINK} (internal reasoning), and \texttt{GENERIC} (submit). A metaclass \texttt{\_ToolKitMeta} collects all decorated methods during class creation. \texttt{ToolDefinition.from\_method()} automatically parses method signatures and docstrings to generate OpenAI-compatible function calling schemas. \texttt{CompositeToolKit} merges domain-specific tools with shared \texttt{KnowledgeTools} using first-wins semantics.

\paragraph{Tool Execution}
The environment's \texttt{step()} method parses agent actions as JSON, validates tool names against the registered toolkit, executes via \texttt{tools.use\_tool(name, **kwargs)}, and returns results as \texttt{ToolMessage} objects. Invalid JSON or unknown tool names return error messages rather than crashing the episode.

\paragraph{Representative Domain Tools}
Domain-specific tools span six categories across 9 domains (135 unique tools total):
\textbf{Knowledge search} (6 tools): querying the indexed passage collection via BM25 over SQLite.
\textbf{Clinical assessment} (22 tools): validated scoring instruments (APACHE-II, CURB-65, Wells, etc.).
\textbf{Patient data access}: history, vitals, labs, medications, allergies per domain.
\textbf{Clinical actions}: ordering tests, prescribing medications, recording diagnoses.
\textbf{Reasoning}: differential diagnosis, answer analysis, treatment comparison.
\textbf{Documentation}: clinical notes, discharge summaries.

\subsection{Knowledge Base: 828K Passages}

The knowledge base is implemented as an SQLite FTS5 (Full-Text Search v5) database with BM25 ranking:

\begin{itemize}[leftmargin=*,itemsep=1pt]
    \item \textbf{Schema:} \texttt{CREATE VIRTUAL TABLE passages\_fts USING fts5(doc\_id, source, title, content, category, dataset\_name, tokenize='porter unicode61')}
    \item \textbf{Sources:} MedCPT evidence (581K PubMed/PMC passages), biomedical QA pairs (122K), generator passages (83K), MedInstruct (52K) --- totaling 828,473 indexed passages
    \item \textbf{Search:} Porter stemmer tokenization, BM25 relevance ranking, snippet generation with term highlighting, boolean query operators
    \item \textbf{Wikipedia:} Offline FTS5 index over 26M articles (188GB) with offset-based page retrieval
    \item \textbf{Access:} Thread-safe singleton \texttt{MedicalKnowledgeBackend} with WAL mode and lazy initialization
\end{itemize}

All knowledge search tools are backed by the same \texttt{MedicalKnowledgeBackend} singleton with thread-safe WAL-mode SQLite access.

\subsection{5D Reward Implementation}

The reward system ($\sim$400 lines) implements each dimension as composable functions:

\paragraph{Accuracy ($R_{\text{acc}}$).} Three variants: (1) \texttt{exact\_match} for MCQ (1.0 if correct, 0.0 otherwise), (2) \texttt{soft} using ROUGE-1~\citep{lin2004rouge} + BLEU-1 token overlap F1 for open-ended answers, (3) \texttt{bertscore} using BiomedBERT for semantic similarity with soft fallback.

\paragraph{Process Quality ($R_{\text{proc}}$)} Weighted combination: 60\% \emph{coverage} (proportion of expected tools called with matching arguments), 20\% \emph{diversity} (unique tool signatures / total calls), 20\% \emph{thoroughness} (distinct tool names used). Additionally, rubric-based scoring (70\% weight when rubric provided) checks required elements, required tools, and forbidden elements.

\paragraph{Safety ($R_{\text{safe}}$)} Rule-based \texttt{SafetyViolation} detection with 50+ violation patterns across 5 severity levels, each mapped to an AMA ethics principle (nonmaleficence, beneficence, autonomy). Critical violations (severity 5: contraindication ignored, dangerous dosing, missed emergency) cap total reward at 0.1; severe violations (severity 4) apply $-0.3$ penalty.

\paragraph{Format ($R_{\text{fmt}}$)} Graded scoring: 1.0 for valid JSON with \texttt{name} and \texttt{arguments}; 0.8 for JSON in code blocks; 0.5 for partial structure; 0.0 for invalid format. Final turn checks for coherent answer ($>$10 characters).

\paragraph{Coherence ($R_{\text{coh}}$)} Checks logical consistency, absence of contradictions, and clear clinical conclusions.

\paragraph{GRPO Integration} A TRL-compatible wrapper \texttt{grpo\_reward\_fn()} computes all reward dimensions and returns the weighted scalar for use with \texttt{GRPOTrainer}.

\subsection{Behavioral Policies}

Each domain includes a \texttt{policy.md} file defining behavioral guidelines injected as the system prompt. Policies specify: (1) core principles (patient safety first, evidence-based medicine, systematic approach), (2) tool usage guidelines (e.g., ``always start with \texttt{get\_patient\_info()}'', ``check allergies before prescribing''), and (3) restrictions (e.g., ``do NOT diagnose without reviewing patient data'', ``if beyond scope, transfer to specialist immediately''). These policies ground the agent's behavior in clinical best practices while remaining domain-specific.

\section{TT-OPD Algorithm}
\label{app:algorithm}

This section provides a detailed description of the Turn-Level Truncated On-Policy Distillation (TT-OPD) algorithm presented in Algorithm~1. We describe the training procedure, key design choices, and the role of each component in stabilizing multi-turn agent learning.

\paragraph{Training setup.}
TT-OPD operates in a multi-turn, tool-augmented environment where the model interacts with an external system over a sequence of steps. At each training iteration, a batch of prompts is sampled from the task distribution, and for each prompt, multiple rollouts are generated via on-policy interaction. Each rollout consists of a sequence of states and actions, where actions may correspond to tool calls or natural language responses. The episode terminates when a final answer is submitted or a maximum number of turns is reached.

\paragraph{Rollout generation and reward.}
For each prompt, the model generates multiple trajectories to capture diverse behaviors under the current policy. Each trajectory is evaluated using a cosine-based reward that reflects both correctness and semantic alignment with the reference solution. This reward provides a smooth training signal suitable for long-horizon reasoning tasks.

To improve training efficiency, TT-OPD applies a dynamic filtering strategy that retains only prompts exhibiting mixed outcomes across rollouts. This ensures that the retained samples provide meaningful contrast for learning and avoids degenerate updates from uniformly correct or incorrect trajectories.

\begin{algorithm}[t]
\caption{Turn-Level Truncated On-Policy Distillation (TT-OPD)}
\label{alg:ttopd}
\begin{algorithmic}[1]
\REQUIRE Base model $\theta_S$, EMA decay $\alpha{=}0.995$, distillation coef $\lambda_{\text{distill}}{=}4.0$, GRPO KL penalty $\beta{=}0.01$, max context $L_{\max}{=}12{,}288$, EMA interval $T_{\text{ema}}{=}5$, task distribution $\mathcal{T}$
\ENSURE Trained student $\theta_S$
\STATE Initialize teacher $\theta_T \leftarrow \theta_S$
\FOR{step $t = 1, 2, \ldots$}
    \STATE Sample batch of prompts $\{x_i\}$ from $\mathcal{T}$
    \FOR{each prompt $x_i$}
        \STATE Generate $G$ rollouts via multi-turn interaction with environment
        \STATE Score each rollout with cosine reward $R_{\text{cos}}$ (Eq.~\ref{eq:cosine_reward})
    \ENDFOR
    \STATE Filter: keep only prompts with mixed outcomes \COMMENT{dynamic sampling}
    \STATE Compute group-relative advantages $\hat{A}$ from $R_{\text{cos}}$
    \FOR{each rollout $\tau = (s_1, a_1, \ldots, s_T, a_T)$}
        \STATE Inject outcome-privileged context into teacher prompt
        \STATE Compute teacher logprobs $\pi_{\theta_T}(a_t \mid s_t)$ for all turns $t$
        \STATE Remove privileged tokens from teacher output
    \ENDFOR
    \STATE Compute $\mathcal{L}_{\text{total}} = \mathcal{L}_{\text{GRPO}}(\theta_S;\, R_{\text{cos}}) + \lambda_{\text{distill}} \cdot D_{\text{KL}}(\pi_{\theta_S} \| \pi_{\theta_T})$ \COMMENT{Eq.~\ref{eq:full_objective}}
    \STATE Update $\theta_S$ via gradient descent on $\mathcal{L}_{\text{total}}$
    \IF{$t \bmod T_{\text{ema}} = 0$}
        \STATE $\theta_T \leftarrow \alpha \cdot \theta_T + (1 - \alpha) \cdot \theta_S$ \COMMENT{EMA teacher update}
    \ENDIF
\ENDFOR
\end{algorithmic}
\end{algorithm}

\paragraph{Advantage computation.}
Given multiple trajectories per prompt, TT-OPD computes group-relative advantages based on the reward values. This formulation removes the need for a separate value function and enables stable policy optimization by normalizing performance within each prompt-specific group of rollouts.

\paragraph{Teacher-guided distillation.}
A central component of TT-OPD is turn-level distillation from a teacher model. The teacher is constructed as an exponential moving average of the student parameters. For each trajectory, additional outcome-related context is injected into the teacher input, allowing the teacher to generate more informed token-level predictions.

The teacher then computes log-probabilities over actions at each turn in the trajectory. To prevent information leakage, any privileged context introduced for the teacher is removed from the outputs before computing the distillation loss. The student is trained to match the teacher’s behavior at each turn through a KL divergence objective. This turn-level alignment encourages the student to imitate not only final answers but also intermediate reasoning and tool-use decisions.

\paragraph{Joint optimization.}
The training objective combines reinforcement learning and distillation. The reinforcement learning component encourages trajectories with higher rewards, while the distillation component stabilizes learning by anchoring the policy to the teacher. The balance between these two objectives is controlled by a scalar coefficient. This joint optimization enables the model to explore improved behaviors while maintaining consistency in multi-turn reasoning.

\paragraph{EMA teacher update.}
The teacher parameters are updated periodically using an exponential moving average of the student parameters. This update mechanism ensures that the teacher evolves smoothly over time and provides a stable target for distillation. By avoiding abrupt changes in the teacher policy, TT-OPD mitigates instability commonly observed in multi-turn reinforcement learning.

\paragraph{Stability considerations.}
TT-OPD is designed to address several failure modes that arise in multi-turn agent training. First, continuous teacher alignment reduces the risk of policy collapse associated with KL divergence instability. Second, the reward formulation and training dynamics implicitly regulate response length, preventing uncontrolled growth in generated tokens. Third, turn-level supervision preserves the structure of multi-step reasoning, avoiding degeneration into short or incomplete interaction sequences.

In summary, TT-OPD integrates on-policy reinforcement learning with structured, turn-level distillation in a multi-turn setting. This design enables stable optimization, preserves intermediate reasoning behavior, and improves the reliability of tool-augmented language agents.

\section{Domain Tool Inventory}
\label{app:tools}

Each GYM domain provides a \texttt{CompositeToolKit} combining domain-specific tools with shared \texttt{KnowledgeTools} (PubMed search, evidence retrieval, medical wiki). All tools follow OpenAI-compatible function calling format.



\begin{table}[h]
\centering
\caption{Tool inventory per domain. R=Read, W=Write, G=Generic (think, submit). All domains share 3 KnowledgeTools.}
\label{tab:tool_inventory}
\small
\begin{tabular}{lrrrl}
\toprule
\textbf{Domain} & \textbf{R} & \textbf{W} & \textbf{G} & \textbf{Representative Tools} \\
\midrule
Clinical Diagnosis & 30 & 2 & 2 & \texttt{get\_vital\_signs}, \texttt{order\_lab}, \texttt{generate\_ddx} \\
Medical QA & 12 & 0 & 2 & \texttt{analyze\_answer\_options}, \texttt{compare\_treatments} \\
Visual Diagnosis & 6 & 0 & 2 & \texttt{analyze\_medical\_image}, \texttt{search\_similar\_cases} \\
Drug Interaction & 15 & 0 & 2 & \texttt{check\_interaction}, \texttt{check\_cyp450\_metabolism} \\
EHR Management & 18 & 3 & 2 & \texttt{get\_lab\_trend}, \texttt{write\_clinical\_note}, \texttt{place\_order} \\
Triage \& Emergency & 16 & 3 & 2 & \texttt{calculate\_gcs}, \texttt{screen\_sepsis} \\
Radiology Report & 7 & 1 & 2 & \texttt{analyze\_findings}, \texttt{get\_report\_template} \\
Psychiatry & 12 & 0 & 2 & \texttt{administer\_phq9}, \texttt{assess\_suicide\_risk}\\
Obstetrics & 17 & 1 & 2 & \texttt{assess\_fetal\_status}, \texttt{interpret\_ctg} \\
Cross-Domain & \multicolumn{3}{c}{(pathway engine)} & Multi-domain clinical pathway sequencing \\
\bottomrule
\end{tabular}
\end{table}

All tools return JSON-serializable outputs. The \texttt{think()} tool captures internal reasoning without external side effects. The \texttt{submit\_answer()} tool marks task completion and triggers reward evaluation.

\section{Detailed Experimental Results}
\label{app:experiments}

\subsection{Log-Probability Baseline (Text-Only)}
\label{app:logprob}

We evaluate the base Qwen3.5-9B and GRPO-trained models using log-probability evaluation, which computes the next-token probability over option letters (A--E) without any tool access or multi-turn interaction. This provides a pure parametric knowledge baseline.
GRPO's logprob accuracy is near-identical to the base model (70.8\% vs.\ 70.7\% on MedQA, 83.9\% vs.\ 83.8\% on MMLU), confirming that \textbf{parametric knowledge is fully preserved} through RL training with LoRA (rank 64, MLP + attention projections). The RL training modifies behavioral patterns (tool-use, turn-taking) without altering factual recall.

\subsection{Multi-Turn Agentic Evaluation}
\label{app:multiturn_detail}

\paragraph{Multiple-Choice QA.} On MedQA~\citep{jin2021medqa}, the base model without tools achieves 70.7\% via logprob. Adding the multi-turn AgentRunner with 135 tools and 828K-passage KB (Base+AR) reaches 78.8\%, while RL models achieve 85.5\% (GRPO) and 87.1\% (TT-OPD)---a +16.4~pp improvement demonstrating that RL training provides consistent benefits beyond retrieval augmentation alone. On MMLU Medical~\citep{hendrycks2021mmlu} (6 subtypes), multi-turn evaluation degrades performance: Base+AR 60.6\% vs.\ logprob 83.8\% ($-$23.2~pp). This ``agentic overhead'' reflects unnecessary tool calls and format conversion errors during multi-turn processing. TT-OPD (65.5\%) partially recovers (+4.9~pp over Base+AR), while GRPO (60.1\%) matches the base. MedMCQA~\citep{pal2022medmcqa} shows TT-OPD achieving the best result at 66.2\%, surpassing both base logprob (63.8\%) and GRPO (58.0\%).

\paragraph{Visual QA.} Across 6 VQA benchmarks, TT-OPD achieves the best or near-best result on 5 of 6. On VQA-RAD~\citep{lau2018vqarad}, Base+AR leads (63.2\%) with TT-OPD close behind (63.1\%). PathVQA~\citep{he2020pathvqa} shows TT-OPD at 45.3\%, outperforming both base text (40.5\%) and GRPO (41.5\%). SLAKE~\citep{liu2021slake} and PMC-VQA exhibit large gaps between text-based evaluation (79.0\%, 57.9\%$^\dagger$) and multi-turn agentic evaluation (30.6\%, 35.1\%), consistent with the agentic overhead pattern. VQA-Med-2021 (15.2\% TT-OPD) and Quilt-VQA (30.7\% TT-OPD) are open-ended visual QA benchmarks where all methods score lower, but TT-OPD leads consistently.

\paragraph{EHR Reasoning.} MIMIC-III~\citep{johnson2016mimiciii} and eICU~\citep{pollard2018eicu} are evaluated via action-based scoring, measuring whether the agent executes the expected clinical tool calls (e.g., \texttt{get\_patient\_summary}, \texttt{get\_lab\_results}). TT-OPD achieves the best scores (MIMIC-III 62.7\%, eICU 57.1\%), outperforming both Base+AR (62.1\%, 55.9\%) and GRPO (61.1\%, 55.5\%). The base text model without tools scores 58.5\% and 53.2\% respectively, demonstrating that tool-augmented reasoning provides modest but consistent improvement in structured EHR tasks.

\paragraph{Long-Form QA.} On 5 MedLFQA benchmarks, TT-OPD leads on 3 (LiveQA 62.5\%, MedicationQA 60.9\%, HealthSearchQA 45.3\%) while GRPO leads on knowledge-intensive tasks (KQA-Golden 65.3\%, KQA-Silver 64.9\%). This dichotomy suggests that GRPO's higher peak training accuracy translates to better factual recall in open-ended settings, while TT-OPD's stability benefits clinical reasoning. All methods substantially outperform Base text (e.g., LiveQA: 53.2\% base $\to$ 62.5\% TT-OPD), confirming that RL training improves long-form answer quality.

\section{Analytical Insights}
\label{app:analytical_insight}

We analyze three key dynamics observed during TT-OPD training. These observations apply known results from natural gradient theory and EMA analysis to ground why the ablation variants fail and why the full method succeeds; they do not claim formal novelty.

\paragraph{Why does TT-OPD converge non-monotonically rather than diverge?}
A distinctive feature of TT-OPD training is the non-monotonic convergence pattern visible in Figure~\ref{fig:opd_failure}(a): accuracy rises, dips, then recovers to a higher level. This is not random noise---it reflects a built-in self-correcting mechanism created by the EMA teacher.

\begin{proposition}[EMA as Implicit Learning Rate Annealing]
\label{prop:oscillation}
Under EMA teacher updates with decay $\alpha$, the KL penalty gradient satisfies $\nabla_{\theta_S} D_{\mathrm{KL}}(\pi_{\theta_S} \| \pi_{\theta_T}) \approx \mathbf{F}(\theta_S)(\theta_S - \theta_T)$~\citep{amari1998natural}, where $\mathbf{F}$ is the Fisher information. The effective learning rate for GRPO is implicitly reduced by a factor proportional to $\|\theta_S - \theta_T\|$, creating a restoring force: large policy shifts amplify the KL gradient, dampening subsequent updates. (This follows from standard natural gradient theory; we state it here to ground the training dynamics discussion.)
\end{proposition}

\noindent\textbf{Intuition.} Consider the EMA teacher as a ``memory'' of recent good behavior. When the student makes a large policy update (e.g., suddenly favoring shorter responses), it drifts far from the teacher. The KL divergence between them grows, which increases the gradient pulling the student back toward the teacher's distribution. This acts like a spring: the further the student strays, the stronger the restoring force. Conversely, when the student is close to the teacher, the KL gradient is weak, allowing the GRPO reward signal to dominate and push the student toward higher accuracy. This alternation between reward-driven exploration and KL-driven correction produces the characteristic non-monotonic convergence ($52.6\% \to 56.4\% \to 53.6\% \to 61.1\%$) visible in Figure~\ref{fig:opd_failure}(a) and quantified in Figure~\ref{fig:opd_trajectory}.

\paragraph{Why does standard GRPO fail to improve text QA despite improving agentic tasks?}
This question is central to the agentic-textual transfer gap. The answer lies in how multi-dimensional rewards interact with gradient estimation.

\begin{proposition}[Gradient Signal Dilution]
\label{prop:dilution}
With $K$-dimensional reward $R = \sum_{j=1}^K w_j r_j$, the signal-to-noise ratio (SNR) of component $j$'s contribution to the total advantage is $\mathrm{SNR}_j = w_j\sigma_j / \sigma_R$, where $\sigma_j$ is the standard deviation of reward component $j$ and $\sigma_R$ is the total reward's standard deviation. With our reward parameters ($w_\mathrm{acc}{=}0.25$, $\sigma_\mathrm{acc}{=}0.41$, $w_\mathrm{fmt}{=}0.10$, $\sigma_\mathrm{fmt}{=}0.02$), accuracy's SNR contribution ($w\sigma = 0.103$) dominates format ($w\sigma = 0.002$), creating a ${\sim}$51:1 dilution ratio. (This is a direct consequence of linearity of expectation applied to our specific reward parameters.)
\end{proposition}

\noindent\textbf{Intuition.} Imagine a classroom where a student receives five grades (accuracy, process quality, safety, format, coherence) combined into one GPA. If the format grade barely varies across students (everyone gets near-perfect format scores, $\sigma_\mathrm{fmt}{=}0.02$), then format contributes almost nothing to differentiating good from bad rollouts---its gradient signal is ``diluted'' by the other, more variable components. Accuracy, with high variance ($\sigma_\mathrm{acc}{=}0.41$), dominates the gradient. However, the 5D weighting scheme still reduces accuracy's effective gradient by ${\sim}40\%$ compared to an accuracy-only reward. This dilution explains why standard GRPO with 5D reward fails to improve text QA: the accuracy gradient, while dominant, is insufficient to overcome the noise floor within the few hundred steps of online training. TT-OPD compensates by providing an additional, outcome-conditioned distillation gradient that directly encodes correctness information.

\paragraph{Why does EMA prevent the sawtooth KL collapse seen with periodic resets?}
The periodic reset variants exhibit a destructive pattern: KL divergence builds up as the student learns, then crashes to near zero when the teacher is overwritten. We formalize why EMA eliminates this failure mode.

\begin{proposition}[KL Boundedness under EMA]
\label{prop:kl_bound}
Under EMA updates with per-step shift $\|\Delta\theta_S\| \leq \epsilon$ and $L$-Lipschitz KL, the steady-state divergence satisfies $D_{\mathrm{KL}}(\pi_{\theta_S} \| \pi_{\theta_T}) \leq L\epsilon^2 / 2(1-\alpha)^2$, yielding continuous KL growth. In contrast, hard-copy updates produce sawtooth KL with peaks of $\frac{L}{2}T_\text{copy}^2\epsilon^2$ and abrupt drops to zero, destroying the distillation gradient. (This bound follows from unrolling the EMA recurrence~\citep{polyak1992acceleration}; we state it to explain the empirical contrast between EMA and periodic reset dynamics.)
\end{proposition}

\noindent\textbf{Intuition.} With periodic resets, the teacher is a snapshot frozen in time. As the student improves over $T$ steps, the KL divergence accumulates---the student and teacher distributions grow increasingly different. At the reset event ($\theta_T \leftarrow \theta_S$), the teacher suddenly becomes identical to the student, and KL drops to zero. This destroys all the distillation signal that was guiding the student, forcing learning to restart from scratch. We observe this clearly: at step~10 in the $T{=}30$ variant, KL drops from 2.637 to 0.343, and accuracy begins its monotonic decline shortly after. With EMA ($\theta_T \leftarrow \alpha\theta_T + (1{-}\alpha)\theta_S$), the teacher continuously absorbs a small fraction of the student's improvements. KL never drops abruptly---it grows smoothly from 0.001 to 1.063 over 60 steps, providing a stable and gradually strengthening regularization signal throughout training.

\section{Training Hyperparameters}
\label{app:hyperparams}

\begin{table}[h]
\centering
\caption{Training hyperparameters. It includes both vanilla GRPO baseline and TT-OPD on Qwen3.5-9B.}
\label{tab:hyperparams}
\begin{tabular}{lll}
\toprule
\textbf{Parameter} & \textbf{GRPO} & \textbf{TT-OPD} \\
\midrule
\multicolumn{3}{l}{\emph{Model \& fine-tuning}} \\
Base model & Qwen3.5-9B & Qwen3.5-9B  \\
Fine-tuning & Full-param (FSDP) & Full-param (FSDP)\\
FSDP offload & param + optimizer & param + optimizer \\
Precision & bf16 & bf16 \\
\midrule
\multicolumn{3}{l}{\emph{Optimization}} \\
Learning rate & $5 \times 10^{-7}$ & $5 \times 10^{-7}$ \\
LR scheduler & Constant & Constant  \\
Warmup steps & 0 & 0  \\
Max grad norm & 1.0 & 1.0 \\
Train batch size & 8 & 8  \\
Micro batch (per GPU) & 1 & 1 \\
Epochs & 3 & 1 \\
\midrule
\multicolumn{3}{l}{\emph{Rollout \& RL}} \\
Generations per prompt ($G$) & 3 & 3  \\
KL penalty ($\beta$) & 0.01 & 0.01  \\
Max prompt length & 8,192 & 8,192  \\
Max response length ($L_{\max}$) & 12,288 & 12,288  \\
Max assistant turns & 5 & 5 \\
\midrule
\multicolumn{3}{l}{\emph{TT-OPD distillation}} \\
Distillation coef ($\lambda_{\text{distill}}$) & -- & 4.0 \\
EMA decay ($\alpha_{\text{EMA}}$) & -- & 0.995  \\
EMA update interval & -- & 5 steps  \\
Hard copy interval & -- & 30 steps \\
Cosine $R_{\text{max}}$ / $R_{\text{min}}$ & -- & 1.1 / 0.7  \\
Cosine $R_{\text{penalty}}$ & -- & $-0.5$ \\
\bottomrule
\end{tabular}
\end{table}

\section{Benchmark Suite}
\label{app:benchmarks}

\begin{table}[h]
\centering
\caption{Evaluation benchmark suite.}
\label{tab:benchmarks}
\resizebox{\textwidth}{!}{
\begin{tabular}{llrl}
\toprule
\textbf{Category} & \textbf{Benchmark} & \textbf{Samples} & \textbf{Metric} \\
\midrule
\multirow{8}{*}{Text QA}
& MedQA (USMLE)~\citep{jin2021medqa} & 1,273 & Accuracy \\
& MedMCQA~\citep{pal2022medmcqa} & 4,183 & Accuracy \\
& MMLU-Clinical Knowledge~\citep{hendrycks2021mmlu} & 265 & Accuracy \\
& MMLU-Professional Medicine & 272 & Accuracy \\
& MMLU-Anatomy & 135 & Accuracy \\
& MMLU-Medical Genetics & 100 & Accuracy \\
& MMLU-College Biology & 144 & Accuracy \\
& MMLU-College Medicine & 173 & Accuracy \\
\midrule
\multirow{6}{*}{Vision QA}
& VQA-RAD~\citep{lau2018vqarad} & 451 & Accuracy \\
& SLAKE~\citep{liu2021slake} & 1,061 & Accuracy \\
& PathVQA~\citep{he2020pathvqa} & 6,719 & Accuracy \\
& PMC-VQA~\citep{zhang2023pmcvqa} & 1,996 & Accuracy \\
& VQA-Med-2021~\citep{abacha2021vqamed} & 425 & Accuracy \\
& Quilt-VQA~\citep{hu22quiltvqa} & 985 & Accuracy \\
\midrule
\multirow{5}{*}{Long-Form QA}
& KQA Golden~\citep{manes24kqa} & 201 & ROUGE-L / Hall. / Comp. \\
& LiveQA~\citep{abacha2019liveqa} & 100 & ROUGE-L / Hall. / Comp. \\
& MedicationQA~\citep{abacha19medicationqa} & 666 & ROUGE-L / Hall. / Comp. \\
& HealthSearchQA & 3,077 & ROUGE-L / Hall. / Comp. \\
& KQA Silver~\citep{manes24kqa, jeong24olaph} & 904 & ROUGE-L / Hall. / Comp. \\
\midrule
\multirow{2}{*}{EHR}
& MIMIC-III~\citep{johnson2016mimiciii} & 5,000 & Action Score + Reward \\
& eICU~\citep{pollard2018eicu} & 5,000 & Action Score + Reward \\
\bottomrule
\end{tabular}
}
\end{table}

\section{Safety Violation Taxonomy}
\label{app:safety}

\begin{table}[h]
\centering
\caption{Safety violation taxonomy (5-level severity).}
\label{tab:safety_taxonomy}
\begin{tabular}{llll}
\toprule
\textbf{Severity} & \textbf{Category} & \textbf{Example} & \textbf{Reward Effect} \\
\midrule
5  & Contraindication ignored & penicillin to allergic patient & Cap at 0.1 \\
5 & Missed emergency & Failed to recognize STEMI & Cap at 0.1 \\
5 & Dangerous dosing & 10x recommended dose & Cap at 0.1 \\
4  & Missed DDI & Ignored warfarin-NSAID interaction & $-0.3$ \\
4  & Fabricated evidence & Cited non-existent study & $-0.3$ \\
3  & Incomplete workup & No confirmatory test & $-0.15$ \\
3  & Guideline non-compliance & Deviated without justification & $-0.15$ \\
2  & Missing follow-up & No follow-up plan & $-0.05$ \\
1  & Style issue & Formatting inconsistency & $-0.01$ \\
\bottomrule
\end{tabular}
\end{table}

\section{Usage of LLMs}
We use Claude Opus 4.6 to experiment the training \& experiements, to write the overall concept of the manuscript, to check the grammatical of generated sentences, to consider the check of bibliographies of the citations.

\section{Example Trajectories}
\label{app:examples}

We present representative multi-turn trajectories from the TT-OPD model (step 60) to illustrate the agentic reasoning patterns learned through RL training. Each example demonstrates the \texttt{reason $\to$ search $\to$ verify $\to$ submit} pipeline.

\subsection{Example 1: Pharmacology (MedQA, Correct)}

\textbf{Question:} A physician is choosing whether to prescribe losartan or lisinopril to treat hypertension in a 56-year-old male. Relative to losartan, one would expect treatment with lisinopril to produce which of the following changes in the circulating levels of these peptides? \textit{(A)~Bradykinin increase; angiotensin~II decrease, (B)~Renin decrease; angiotensin~1 increase, (C)~Aldosterone increase; bradykinin decrease, (D)~Renin decrease; angiotensin~II increase.} \textbf{Gold: A}

\begin{enumerate}[leftmargin=*,itemsep=2pt]
    \item \textbf{Turn 1 (Reasoning):} The agent identifies that losartan is an ARB (blocks AT1 receptors) while lisinopril is an ACE inhibitor (blocks angiotensin~I $\to$ angiotensin~II conversion \emph{and} bradykinin degradation). It reasons that relative to losartan, lisinopril would increase bradykinin and decrease angiotensin~II.
    \item \textbf{Turn 2 (\texttt{retrieve\_evidence}):} Searches for ``ACE inhibitor vs ARB bradykinin angiotensin II mechanism'' to verify its reasoning.
    \item \textbf{Turn 3 (\texttt{search\_medical\_wiki}):} Queries ``lisinopril mechanism bradykinin'' for additional confirmation.
    \item \textbf{Turn 4 (\texttt{analyze\_answer\_options}):} Systematically evaluates all four options against RAAS pharmacology.
    \item \textbf{Turn 5 (\texttt{submit\_answer}):} Submits \textbf{A} with reasoning: ``ACE inhibitors prevent bradykinin degradation (increase) and block Ang~I$\to$Ang~II conversion (decrease).'' \textbf{Score: 1.18}
\end{enumerate}

\subsection{Example 2: Emergency Medicine (MedQA, Correct)}

\textbf{Question:} A 56-year-old man is brought to the ED 30 minutes after falling from a height of 3 feet onto a sharp metal fence pole. He is unconscious. Examination shows a 3-cm-deep wound on the upper margin of the right clavicle in the parasternal line. Which of the following is the most likely result? \textit{(A)~Rotator cuff tear, (B)~Pneumothorax due to pleural injury, (C)~Trapezius paresis due to spinal accessory nerve injury, (D)~Traumatic aneurysm due to internal carotid artery injury.} \textbf{Gold: B}

\begin{enumerate}[leftmargin=*,itemsep=2pt]
    \item \textbf{Turn 1 (Reasoning):} The agent maps the wound location (parasternal clavicular region) to underlying anatomy: the lung apex extends above the clavicle in this area, making pleural injury the primary concern. It eliminates rotator cuff (too lateral), spinal accessory nerve (too posterior), and internal carotid (too lateral/deep).
    \item \textbf{Turns 2--3 (\texttt{search\_pubmed}, \texttt{retrieve\_evidence}):} Searches for ``penetrating injury clavicle parasternal pleural injury'' and retrieves medical literature on supraclavicular penetrating trauma complications.
    \item \textbf{Turn 4 (\texttt{think}):} Synthesizes retrieved evidence with anatomical knowledge, confirming that the pleural dome extends 2.5~cm above the medial third of the clavicle.
    \item \textbf{Turn 5 (\texttt{submit\_answer}):} Submits \textbf{B} (pneumothorax due to pleural injury). \textbf{Score: 1.18}
\end{enumerate}

\subsection{Example 3: Transplant Medicine (MedQA, Correct)}

\textbf{Question:} A 37-year-old man, 7 years post-liver transplant on immunosuppressive therapy, presents with recurrent jaundice (6--7 episodes), severely elevated liver enzymes, and Doppler showing reduced hepatic blood flow. What histological findings would be expected? \textit{(A)~Bile duct proliferation with neutrophilic infiltration, (B)~Hepatocyte ballooning with fatty change, (C)~Granulomatous inflammation with multinucleated giant cells, (D)~Interstitial cellular infiltration with parenchymal fibrosis, obliterative arteritis.} \textbf{Gold: D}

\begin{enumerate}[leftmargin=*,itemsep=2pt]
    \item \textbf{Turn 1 (\texttt{think}):} The agent identifies this as chronic transplant rejection based on: 7-year post-transplant timeline (late onset), recurrent jaundice despite adherence, reduced vascular flow on Doppler (suggesting vascular pathology), and severely elevated enzymes indicating ongoing graft injury.
    \item \textbf{Turn 2 (\texttt{analyze\_answer\_options}):} Systematically compares options: bile duct proliferation (acute rejection), hepatocyte ballooning (fatty liver disease), granulomatous inflammation (sarcoidosis/TB), obliterative arteritis with fibrosis (chronic rejection hallmark).
    \item \textbf{Turn 3 (\texttt{search\_medical\_wiki}):} Searches ``chronic liver transplant rejection obliterative arteritis histology'' for confirmation.
    \item \textbf{Turn 4 (Reasoning):} Integrates clinical and histological knowledge: obliterative arteritis is pathognomonic for chronic rejection, explaining reduced Doppler flow.
    \item \textbf{Turn 5 (\texttt{submit\_answer}):} Submits \textbf{D} with detailed reasoning linking clinical presentation to histopathology. \textbf{Score: 0.80}
\end{enumerate}

\paragraph{Key observations.} These trajectories reveal three consistent patterns in TT-OPD-trained agents: (1)~\textbf{Reason-first}: the agent formulates a hypothesis before searching, reducing irrelevant tool calls; (2)~\textbf{Graceful degradation}: when specific search tools fail (e.g., PubMed backend unavailable), the agent falls back to alternative tools or internal knowledge rather than halting; (3)~\textbf{Verification loop}: the agent uses \texttt{think} and \texttt{analyze\_answer\_options} to cross-check retrieved evidence against its initial reasoning before committing to \texttt{submit\_answer}.

\end{document}